# Human evaluation of robotic grippers for berry picking

Laura Álvarez-Hidalgo & Ian S. Howard

SECAM, University of Plymouth, PL4 8AA, Plymouth
laura.alvarezhidalgo@students.plymouth.ac.uk, ian.howard@plymouth.ac.uk

**Abstract.** We describe the construction and evaluation of two robotic grippers for berry picking. Using a pneumatic cylinder drive, one was constructed from hard materials and the other from soft materials. A novel evaluation paradigm using a handle mechanism was developed, so the grippers could be directly operated by human participants. An artificial bush was also constructed and used for evaluation purposes. Overall, both grippers performed worse than the human hand, indicating that further development is needed.

## 1 Introduction

There is a growing interest in robotic harvesting technologies [1, 2]. Recently there has also been significant development in the field of 3D printed robot components for harvesting, including the use of soft grippers [3], which are an area of intense research [4] and offer benefits such as low cost and ease of manufacture [5]. Here we investigate soft and hard grippers for integration in low-cost agricultural robot systems. We use human evaluation using the grippers as tools in picking tasks, thereby decoupling the operation of potential AI controllers from the intrinsic capabilities of each gripper design.

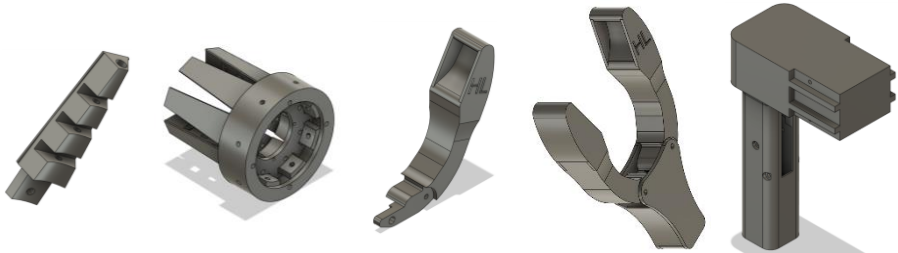

**Fig. 1.** Soft Gripper Finger and Hand. **Fig. 2.** Hard Gripper Finger and Hand. **Fig. 3.** Handle.

## 2 Mechanism Design

We first developed a soft gripper with eight Ninja flex fingers, which can flex and grasp objects when its tendons are tensioned. To allow the fingers to bend, each consisted of a continuous back structure with four protruding block sections, separated by gaps (see Fig. 1 Finger). A string tendon was inserted through a small hole in each finger block all the way up to the fingertip, where it was firmly secured by a knot in the end of the string and glued to secure it in place. Pulling the tendon closed the gaps between the blocks and caused the finger to bend. A small location hole at the lower end of each finger enables the attachment of all eight fingers to a circular hand structure using small screws (Fig. 1. Hand). A circular plate was fastened to the end of the pneumatic cylinder and all tendons were attached to it, and pulling the plate closed all eight fingers simultaneously. See Fig. 4A to see the principle of operation. When tendon tension was released, the springiness of the Ninja Flex allowed all fingers to return to their original resting position without the need for active retraction.

The second gripper design used hard materials (Fig. 2). It comprises a pair of PLA+ fingers that were driven open and closed by an air cylinder acting on the end of a parallelogram mechanism (the principle of operation shown in Fig. 5A). To ensure the gripper could effectively grasp berries, both its fingers had a curved opening at their tips designed to accommodate an average sized berry.

To test the grippers, a PLA+ handle was developed (Fig. 3), which enabled human participants to operate them as tools in picking tasks (Figs. 4B & 5B). The handle has a diameter of 33mm to ensure that it could comfortably fit a human operator's hand. An internal microswitch was housed within the handle and used to activate a pneumatic actuator. The handle was hollow and the control cables from the switch to the Arduino Nano exited through its base. Its top provided a gripper attachment point, to ensure interchange between them was quick and easy.

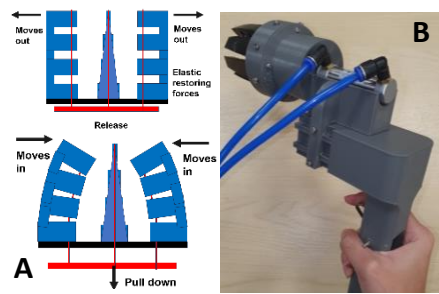

**Fig. 4.** Soft Gripper. **Panel A** Principle of Operation. **Panel B** Gripper Attached to Handle.

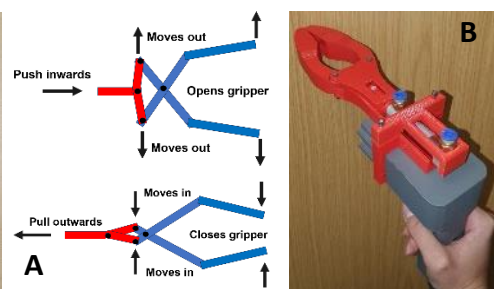

**Fig. 5.** Hard Gripper. **Panel A** Principle of Operation. **Panel B** Gripper Attached to Handle.

A Heschen CDJ2B 16-25 slim, lightweight (20g) double-acting pneumatic cylinders was chosen to actively close the gripper mechanisms (and also open the hard gripper), since it provides a clean and compact form of actuation

[6]. The cylinder had a 16mm bore and 25mm stroke. Running at 5 bar air pressure it could generate up to 50N force. It was controlled using a Heschen 4V210 5-way 2-position 24-volt solenoid valve, which enabled the cylinder rod to be actively driven in and out. Quickfit 6mm ports and 6mm o/d polyethylene tubes were used to connect the air supply, which was provided by a small Clarke air Shhh 50/24 silent air compressor. Both grippers were operated using a switch on the handle mechanism that activated the solenoid valve using a driver circuit implemented on an Arduino Nano (Fig. 6).

To evaluate the grippers' ability to pick berries, a test bush was constructed from a wooden rack (Fig. 7A). Artificial raspberries were hung using Ninja Flex stalks, which released the berry when pulled with sufficient force. This provided a controlled simulation of real berry picking. To achieve this, the berry stem had a cone-like shape that fitted into holes at the base of the berry. The stems themselves were attached to the rack using a simple screw fastening mechanism, allowing for appropriate placement to mimic their natural growth See Fig. 7B to see the 3D printed berry stems.

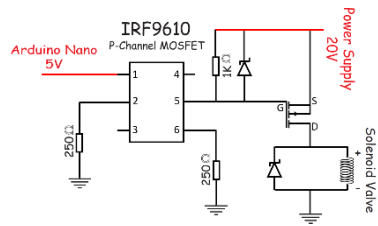
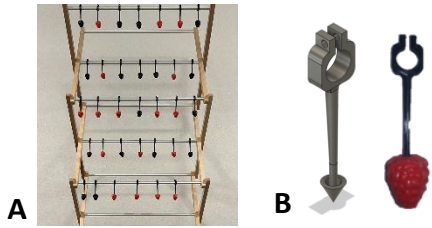

**Fig. 6.** Circuit for Pneumatic Controller     **Fig. 7.** Test Berries and Bush

## 3     Results

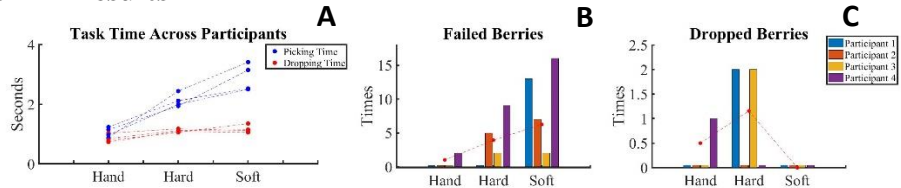

**Fig. 8.** Experimental Results for each Gripper Mechanism. Panel A: Average Picking and Release Time across Participants. Panel B: Berries Missed Panel C: Berries Dropped.

Four participants performed timed experiments to evaluate the effectiveness of using a hard and soft gripper, as well as their hands, in a picking task on an artificial berry bush. Participants were required to pull-off 34 artificial raspberries and collect them in a bowl. A video link showing the picking task is provided **here:**
https://youtube.com/playlist?list=PLFlgfzQylyK69KlPiWzTve7EWyr2Um5gP

We found the time taken to pick a berry varied substantially between conditions, as shown in the plots of the results in Fig. 8. The hand took an average of 1.0s to pick a berry, whereas it took 2.1s and 2.9s with the hard and soft grippers, respectively. Although the soft gripper took almost twice the time compared to the hard gripper to pick the berry, the time advantage was reduced considerably when berry dropping was considered. Results also showed that using the soft gripper, participants tended to miss the berry more often, but once picked, it had a lower chance of being dropped. Conversely, using the hard gripper, participants tended to be more successful in picking a berry but dropped it more often during transfer to the bowl.

## 4 Discussion

Using human evaluation, we demonstrated that two robotic grippers could successfully harvest berries from a test raspberry bush. A hard gripper performed better in terms of berry grasping, whereas a soft gripper exhibited fewer berry dropping errors. Both were worse than using the human hand directly. Clearly, the gripper designs need to be improved, and future experiments are needed to further compare harvesting performance. We suggest that tracking and video of the picking task performed by human operation may provide a useful dataset to train future autonomous robotic systems. Finally, we thank the University of Plymouth for Proof-of-Concept support.